\documentclass{article}
\usepackage{ijcai25}

\usepackage{times}
\usepackage{soul}
\usepackage{url}
\usepackage[hidelinks]{hyperref}
\usepackage[utf8]{inputenc}
\usepackage[small]{caption}
\usepackage{graphicx}
\usepackage{amsmath}
\usepackage{amsthm}
\usepackage{booktabs}
\usepackage{algorithm}
\usepackage{algorithmic}
\usepackage[switch]{lineno}
\usepackage{pgfplots}
\pgfplotsset{compat=1.7}

\usepackage{amssymb}
\usepackage{xcolor}
\usepackage{todonotes}
\usepackage{pgfplots}
\usepackage{booktabs}

\newtheorem{example}{Example}
\newtheorem{definition}{Definition}

\newcommand{\true}{\ensuremath{1}}
\newcommand{\false}{\ensuremath{0}}

\newcommand{\kb}{\ensuremath{\mathcal{K}}}

\newcommand{\atoms}{\ensuremath{\mathsf{At}}}
\newcommand{\allkbs}{\ensuremath{\mathbb{K}}}

\newcommand{\MI}{\textsf{MI}}

\newcommand{\Problematic}{\textsf{Problematic}}

\newcommand{\lang}{\ensuremath{\mathcal{L}}}

\newcommand{\inc}{\ensuremath{\mathcal{I}}}

\newcommand{\modelSet}[1]{\ensuremath{\mathsf{Mod}(#1)}}

\newcommand{\incmi}{\ensuremath{\mathcal{I}_{\mathsf{MI}}}}

\newcommand{\incmv}{\ensuremath{\mathcal{I}_{mv}}}
\newcommand{\incat}{\ensuremath{\mathcal{I}_{at}}}

\title{(Neural-Symbolic) Machine Learning for Inconsistency Measurement}

\author{
Sven Weinzierl$^1$
\and
Carl Corea$^2$\and
\affiliations
$^1$Friedrich-Alexander Universität Erlangen-Nürnberg, Nuremberg, Germany\\
$^2$University of Koblenz, Koblenz, Germany\\
\emails
sven.weinzierl@fau.de,
	ccorea@uni-koblenz.de
}

\begin{document}

\maketitle

\begin{abstract}
We present machine-learning-based approaches for determining the \emph{degree} of inconsistency---which is a numerical value---for propositional logic knowledge bases. Specifically, we present regression- and neural-based models that learn to predict the values that the inconsistency measures $\incmi$ and $\incat$ would assign to propositional logic knowledge bases. Our main motivation is that computing these values conventionally can be hard complexity-wise. As an important addition, we use specific postulates, that is, properties, of the underlying inconsistency measures to infer symbolic rules, which we combine with the learning-based models in the form of constraints. We perform various experiments and show that a) predicting the degree values is feasible in many situations, and b) including the symbolic constraints deduced from the rationality postulates increases the prediction quality.
\end{abstract}

%
%
\section{Introduction}\label{sec:introduction}
Inconsistency in information is a core problem in many AI and knowledge representation tasks. Unfortunately, while there are many results on how to measure and analyze inconsistency (see \cite{thimm2019inconsistency}), computing such results on a \emph{symbolic} level can be hard complexity-wise.
In this work, we therefore present \emph{machine learning}-based approaches for  inconsistency measurement, specifically, predicting the \emph{degree} of inconsistency w.r.t. inconsistency measures. This allows---after training---to obtain degree values for specific knowledge bases in constant time.

An inconsistency measure is any measure that can quantify the \emph{severity} of inconsistency in information with a numerical value. In this work, we refer to inconsistency measures for propositional logic knowledge bases. As an example, consider the knowledge bases $\kb_1, \kb_2$, defined via
\begin{align*}
    \kb_1 = \{a, \neg a\}, && \kb_2 = \{a, \neg a, a\rightarrow b, \neg b\wedge c\}.
\end{align*}
Obviously, both knowledge bases are inconsistent, however, the granularity of the inconsistency is arguably different: For $\kb_1$, there is only one conflict $\{a, \neg a\}$, but for $\kb_2$ we have two conflicts, namely $\{a, \neg a\}$ and $\{a, a\rightarrow b, \neg b\wedge c\}$. To quantify the degree of inconsistency, various measures have been proposed, one of them being the $\incmi$ measure, which counts the number of ``minimal inconsistent subsets" (see conflicting sets above). So here we have that $\incmi(\kb_1)=1$ and $\incmi(\kb_2)=2$. 
In this sense, the inconsistency in $\kb_2$ can be seen as ``more severe" than that in $\kb_1$.

\begin{figure}[H]
    \centering
    \includegraphics[width=1\columnwidth]{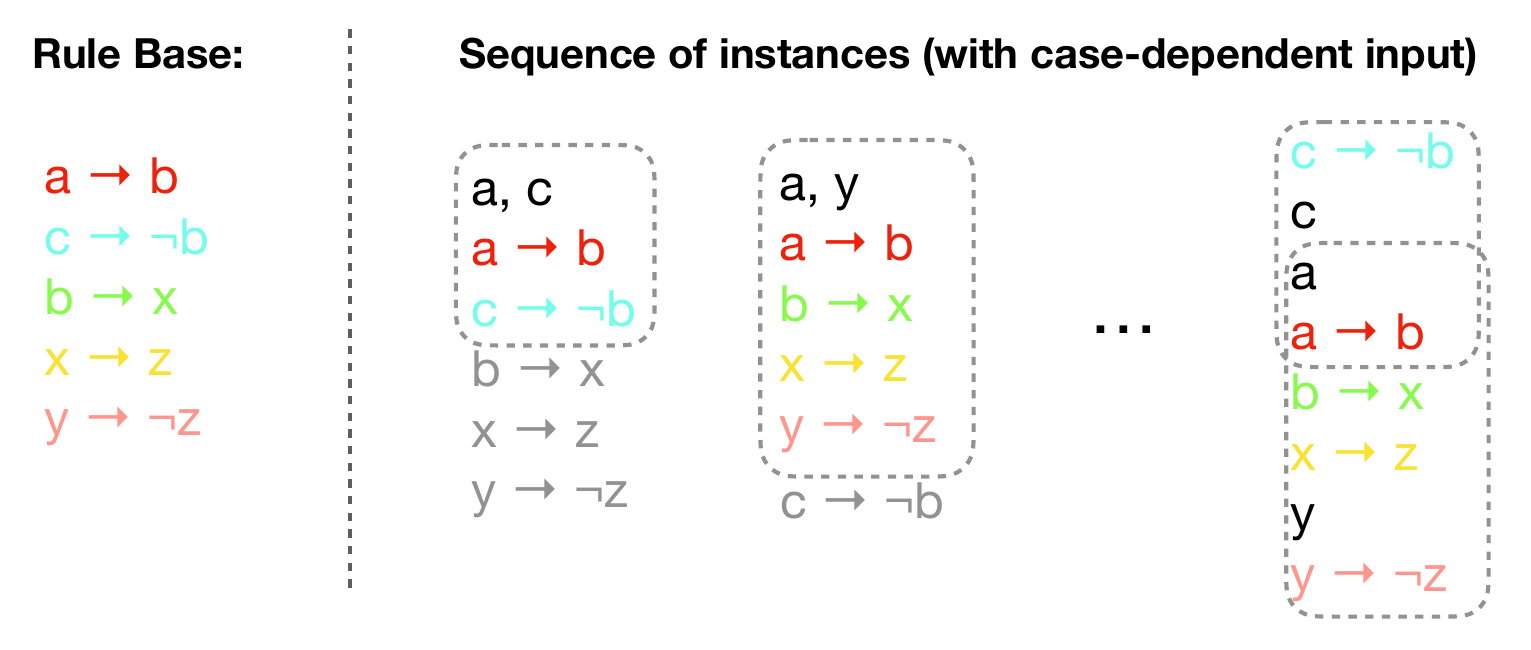}
    \caption{Exemplary instances (with minimal inconsistent subsets), constructed over a rule base and a seq. of case-dependent inputs.}
    \label{fig:exemplaryInstances}
\end{figure}

Inconsistency measurement has gained a nice momentum, however, a problem for application is that many measures are hard to compute computationally \cite{thimm2019complexity}. 
This is especially problematic for settings where a high number of instances have to be assessed \emph{continuously}, as shown in Figure \ref{fig:exemplaryInstances}. In domains such as the financial industry or fraud management, it is common that a shared set of business rules is evaluated against high amounts of case-dependent input (e.g., think of a customer loan application).  

As shown in the example, various different inconsistencies can arise in the instances, depending on the specific input. A problem for such settings is that thousands of cases may have to be analyzed by the company (see \cite{corea2021measuring}). Here, the high computational complexity of solving every instance traditionally may be a potential problem. In response, we propose to use machine learning-based approaches that can learn to \emph{predict} the corresponding inconsistency degree values, based on some instances that have already been solved. This would allow companies to obtain an approximation of the inconsistency values (for future continuous case analysis) in constant time.

For our use-case, we implement regression-based and neural-based models and conduct experiments to investigate the feasibility of our proposed approach. Importantly, we also leverage so-called rationality postulates to infer domain knowledge. This domain knowledge can then be combined with the models as symbolic constraints. We implement and evaluate different architectures for approaching the proposed machine learning problem in general, and for integrating the symbolic constraints. Our experiments show that a) the trained machine learning models can provide useful approximations in many situations, and b) including the symbolic constraints deduced from the rationality postulates increases the prediction performance.

The remainder of this work is structured as follows. We provide preliminaries in Section \ref{sec:preliminaries} and frame the concrete {problem setup} in Section \ref{sec:problemSetup}. We propose several learning-based {architectures} in Section \ref{sec:model} and {evaluate} the proposed approaches in Sections \ref{sec:experimentalSetup} and \ref{sec:results}, where the latter also discusses important countermeasures that were implemented against threats to validity, such as overfitting. Finally, we discuss {scalability} in Section \ref{sec:scalability}, and {conclude} in Section~\ref{sec:conclusion}.

%
%
\section{Preliminaries}\label{sec:preliminaries}

Let $\atoms$ be some fixed set of propositions, and let $\lang(\atoms)$ be a corresponding propositional language constructed using the usual connectives $\wedge$ (\emph{conjunction}), $\vee$ (\emph{disjunction}), 
and $\neg$ (\emph{negation}). A literal is a proposition $p$ or its negation $\neg p$. 

\begin{definition}[Knowledge Base (KB)]\label{def:kb}
	A knowledge base $\kb$ is a finite set of formulas $\kb\subseteq\lang(\atoms)$. Let $\allkbs$ be the set of all knowledge bases.
\end{definition}
For a KB $X$, we denote the propositions in $X$ by $\atoms(X)$.

The semantics for a propositional language is given by \emph{interpretations} where an interpretation $\omega$ on \atoms\ is a function $\omega:\atoms\rightarrow\{\false,\true\}$ (where $0$ stands for false and $1$ stands for true). Let $\Omega(\atoms)$ denote the set of all interpretations for \atoms. An interpretation $\omega$ \emph{satisfies} (or is a \emph{model} of) an atom $a\in\atoms$, denoted by $\omega\models a$, if and only if $\omega(a)=\true$. The satisfaction relation $\models$ is extended to formulas in the usual way. For $\Phi\subseteq\lang(\atoms)$ we also define $\omega\models \Phi$ if and only if $\omega\models\phi$ for every $\phi\in\Phi$.

For every set of formulas $X$, we denote the set of models as 
$\modelSet{X}=\{\omega\in \Omega(\atoms)\mid \omega\models X\}$. 
If $\modelSet{X}=\emptyset$, we write $X\models \perp$ and say that $X$ is \emph{inconsistent}. A set $X$ that is inconsistent and minimal in terms of set inclusion is called minimally inconsistent.


\begin{definition}[Minimal Inconsistent Subset]
    Let a knowledge base $\kb$, a set $M\subseteq\kb$ is called a minimal inconsistent subset (\textsf{MIS}) of $\kb$ if $M\models \perp$ and there is no $M'\subset M$ with $M'\models \perp$. Let $\MI(\kb)$ be the set of all \textsf{MIS}s of $\kb$.
\end{definition}

Any formula $\alpha\in\kb$ that appears in at least one \textsf{MIS} of $\kb$ is called \emph{problematic} (denote $\Problematic(\kb)$ as the set of all problematic formulas in $\kb$).
\begin{example}
   Let  $\kb = \{a, b, \neg b, c, \neg c, d \vee e\}$ then 
$\MI(\kb) = \{ \{b, \neg b\}, \{c, \neg c\}\}$, and $\Problematic(\kb) = \{b, \neg b, c, \neg c\}$. 
\end{example}

For quantifying the degree of inconsistency, we consider inconsistency measures $\inc$, which are functions that assign to a knowledge base a non-negative numerical value (a higher value referring to a higher degree of inconsistency).
Inconsistency measures are commonly divided into \emph{formula-centric measures}, and \emph{atom-centic measures}, where the former focus on individual formulas, and the latter are defined over the signature (propositional atoms) of the knowledge base \cite{hunter2005approaches}. Examples of concrete measures of these two groups are the $\incmi$ measure (formula-centric), and the $\incat$ measure (atom-centric), respectively, which are defined as follows (adapted from \cite{grant2021measuring}).
\begin{definition}[Considered Measures]
    Let $\kb$ be a knowledge base, then define the following two inconsistency measures:
    \begin{itemize}
        \item $\incmi(\kb)=|\MI(\kb)|$
        \item $\incat(\kb)=|\atoms(\Problematic(\kb))|$
    \end{itemize}
\end{definition}
Many other measures have been proposed, see \cite{thimm2019inconsistency}. For the remainder of this work, we will, however, only focus on the two introduced measures $\incmi$ and $\incat$, as they are prominent representatives of formula- and atom-centric measures and can therefore provide insights for approximating measures of these two groups. Also, note that the computation of $\incat$ and $\incmi$ is on the second, resp., above the third, level of the polynomial hierarchy \cite{thimm2019complexity}, so these measures represent good candidates for assessing whether a machine learning-based approach can yield significant performance boosts. 

A final remark on inconsistency measures is that many properties of these measures have been studied (see e.g. \cite{thimm2018evaluation}).
These properties describe the general behavior of the measures, respectively, their values -- for example, under which conditions a measure will return 0, or what possible values the measure can attain.
We will revisit some of these properties in Section \ref{subsec:rules}
and show how the yielded domain knowledge can be leveraged to impose symbolic constraints for the---otherwise sub-symbolic---models.


%
%
\section{Problem Setup}\label{sec:problemSetup}
To compute inconsistency degree values with a machine learning-based approach, we pose the problem of computing these values as a regression problem. 
For this, let $\kb$ be a knowledge base and $\inc$ be an inconsistency measure. Then, our goal is to approximate a function $f_\inc$ mapping the knowledge base $\kb$ to a numerical value $f_\inc(\kb)$, which denotes the inconsistency degree value of $\kb$ w.r.t. $\inc$. 

For representing knowledge bases, we turn to an encoding-based approach. For this, let $\lang({\atoms})$ be a propositional language over $\atoms$ as before, and let $N = N_1,...,N_n$ be a sequence of distinct formulas from $\lang({\atoms})$. Then, for an individual knowledge base $\kb$ and a sequence of words $N$ (both over $\lang({\atoms})$), an encoding can be defined by considering an $|N|-$dimensional vector $V_{\kb}$ s.t.
\begin{align*}
V_\kb[j] = \left\{
\begin{array}{ll}
1 & \text{if \, } N_j\in\kb, \\
0 & \text{otherwise.} \\
\end{array}
\right.
\end{align*}
We investigate the scalability of this encoding-based technique in Section \ref{sec:scalability}.

%
%
\section{Machine Learning Models}\label{sec:model}

The machine learning models presented in this paper are (traditional) regression-based models and neural-based models. They are learned in a supervised way with machine learning algorithms on data sets, where the data consists of knowledge bases represented as feature vectors, and the label is the inconsistency measure $\incmi$ or~$\incat$. 

We differ between four machine learning models. 
The first three models are regression-based and built using the machine learning algorithms linear regression (\textsf{LR}), ridge regression (\textsf{Ridge}), and lasso regression (\textsf{Lasso}). The linear regression models the linear relationship between a label (dependent variable) and the features of the feature vectors (independent variables). In doing so, the effect of each feature on the model output is represented by a coefficient. The ridge regression is an extension of the linear regression and adds a penalty on the size of coefficients to prevent overfitting. The lasso regression reduces overfitting but can also perform feature selection by shrinking coefficients to zero.  

For all three regression-based models, we differ between two variants: i) the feature vectors \emph{only} including features describing the propositional logic formulas, and ii) the feature vectors as in i) $+$ additional features describing (symbolic) domain knowledge (see Section \ref{subsec:rules}). To clarify, these additional features are binary features encoding some external property, where the value 0/1 indicates whether this property holds for the KB instance. We refer to the variants i/ii) as \emph{without/with} flags.    

The last model is built with a multi-layer perceptron (\textsf{MLP}). The MLP consists of three fully-connected (dense) layers. The first layer is the input layer, which maps the input to a hidden state. The second one is a hidden layer and maps the hidden state from the previous layer to a more abstract hidden state. The last layer maps the hidden state of the previous layer to the final prediction. In addition, a $\mathit{tanh}$ activation function is applied to the output of the first layer and the second layer to model non-linearities.  

For this model, we differ between three variants. The first two are similar to the variants of the regression-based models, that is, i) only the KB data without flags, and ii) KB data with additional features encoding domain knowledge.
The third variant is an extension of the second one and considers symbolic domain knowledge in two respects: i) via flags in the feature vectors, and ii) via external constraints in a customized loss function. The customized loss $\mathfrak{L}^{*}$ used in our MLP models is formalized as

\begin{equation}
    \mathfrak{L}^{*} =
    \underbrace{
    \mathfrak{L}_{pred}}_{Prediction \text{ } loss} + 
    \underbrace{
    \mathfrak{L}_{hr,1} + \cdots +
    \mathfrak{L}_{hr,i} + \cdots +
    \mathfrak{L}_{hr,N}}_{(Symbolic) \text{ } domain \text{ } knowledge \text{ } loss}, 
\end{equation}

where $\mathfrak{L}_{pred}$ is the prediction loss calculated as L1 loss, $N$ is the number of heuristics (i.e., external constraints), and $\mathfrak{L}_{hr,i}$ is the loss for the $i$-th heuristic of the (symbolic) domain knowledge. 
$\mathfrak{L}_{hr,i}$ is defined by

\begin{equation}
\mathfrak{L}_{pred} * \overline{x}_{hr,i},      
\end{equation}

where $\overline{x}_{hr,i}$ is the arithmetic mean of the binary values of the feature $x_{hr,i}$, describing the $i$-th heuristic of the (symbolic) domain knowledge. 

Further, if a heuristic $i$ is described by $J$ features, first,   
\begin{equation}
\mathfrak{L}_{pred} * \overline{x}_{hr,i}^{j},
\end{equation}

is calculated per feature $j$, denoting $\mathfrak{L}_{hr,i}^{j}$, and then 

\begin{equation}
\sum_{j=1}^{J}\mathfrak{L}_{hr,i}^{j},
\end{equation}

is calculated, denoting the sum of the losses over the $J$ features describing the $i$-th heuristic.

In general, since the goal of the training procedure is to find the best values of model parameters (i.e., biases and weights) that minimize the custom loss function, the model parameters are also adjusted depending on external constraints defined in the custom loss function.  

To clarify, the second and third variants of the MLP can be considered a neuro-symbolic approach as here symbolic knowledge is used for building neural network models \cite{kautz2022nesy}. The difference is how the domain knowledge is integrated. That is, for variant ii), integrated into the feature vector of the data, and for variant iii), integrated into the model via the custom loss function. 

In summary, the considere models are as follows:
\begin{enumerate}
    \item \textsf{LR} \emph{(with/without flags in feature vector)}
    \item \textsf{Ridge} \emph{(with/without flags in feature vector)}
    \item \textsf{Lasso} \emph{(with/without flags in feature vector)}
    \item \textsf{MLP} \emph{(with/without flags in feature vector; with flags in feature vector and constraints in loss function)}
\end{enumerate}

%
%
\section{Experimental Setup}\label{sec:experimentalSetup}
In this section, we continue to describe the data set generation, the applied symbolic rules/heuristics (``flags"), and the training configuration of our experiments.\footnote{The generated data sets and the developed source code can be found here: \url{https://github.com/fau-is/MLXIncon}}

\subsection{Data Set Generation}\label{subsec:data}
We implemented a generator for creating synthetic knowledge bases. Importantly, the  \textsf{Tweety}-library\footnote{\url{https://tweetyproject.org/index.html}} could be used for computing the inconsistency measures with a conventional solver-based approach (the exact solver is discussed in Section \ref{sec:exp_results}).  So for all knowledge bases, we have the actual value of $\incmi/\text{ }\incat$ as ground truth in the data set (i.e., a triple $\langle\kb,\incmi(\kb),\incmv(\kb)\rangle)$. For knowledge base creation w.r.t. a set of atoms $\atoms$, formulas $\varphi$ 
 of the knowledge base were created via the following syntax:
\begin{align}\label{eq:syntax}
    \varphi ::= l \mid \varphi_1\vee \varphi_2 \mid \varphi_1 \wedge \varphi_2
\end{align}
where $l$ is a literal (i.e., a simple atom or its negated form) over $\atoms$. In other words, we consider a general propositional logic where every formula consists of an arbitrary amount of literals that can be combined arbitrarily with $\vee/\wedge$.
For generating the knowledge bases, we considered as parameters the number of distinct atoms in $\atoms$ from $\{3, 6, 9\}$ and the number of distinct formulas per knowledge base from $\{\leq5,\leq10,\leq15\}$. As a design choice, every formula had at most 10 con-/disjuncts.
\begin{example}
    For a setting of $9$ (atoms) and $\leq15$ (formulas), an exemplary KB yielded by the generator for our experiments was $\kb=\{a \vee (c \wedge d \wedge \neg g) \vee \neg g; 
    a \wedge b \wedge d \wedge \neg f \wedge h; 
    \neg b \wedge f \wedge \neg g \wedge h; 
    \neg b \wedge \neg h \wedge i; 
    (\neg a \wedge b \wedge d \wedge g) \vee h; 
    \neg f \wedge g\}$, where 
    $\incmi(\kb)=5$. 
\end{example}
For each such parameter combination (3 atom settings $\times$ 3 formula settings = 9 combinations), we created individual data sets each containing 1,000 knowledge base instances. The data sets can be accessed online (see above). 
As mentioned, for every knowledge base instance, the actual value of $\incmi/\incat$ was computed using the \textsf{Tweety}-library. 

Table \ref{tab:characteristics} shows the characteristics of the generated data sets. 
%
Some observations follow. First, we observe that the actual number of inconsistencies increases if the number of formulas is increased. Especially for the MI measure, this means that a potentially higher range of target values has to be predicted. 
Furthermore, the \emph{entropy}\footnote{
Entropy is a measure of the ``spread" of values that an inconsistency measure assigns to a sequence of knowledge bases \cite{thimm2019experimental}.
Let an inconsistency measure $\inc$ and a sequence of knowledge bases $K=\kb_1,...,\kb_n$, and denote $I=\inc(\kb_1),...,\inc(\kb_n)$ as a sequence of values $\inc(\kb_i)$ over all elements of $K$; with $I^{U}$ being the set of unique values in $I$. Then, the entropy of all measure values $H(K,\inc)$ is defined as $H(K,\inc)=-\sum_{v\in I^{U}} \frac{f_v}{|K|} \mathit{ln}\frac{f_v}{|K|}$, where $f_v$ is the frequency of $v$ in $I$ and $\mathit{ln}$ is the natural logarithm.} 
of all inconsistency degree values differs for the different settings. 
This allows to test the prediction models for different distributions of target values. 
Some of these instances could also correctly be flagged as consistent with what we call a heuristic (see the following Section). 
%

\begin{table}[H]
    \resizebox{\columnwidth}{!}{
    \begin{tabular}{crrrrr}
        \toprule        
        &  $\inc_{\mathit{MI}}^{\mathit{max}}$  & $\inc_{\mathit{MI}}^{\mathit{ent}}$ &$\inc_{\mathit{at}}^{\mathit{max}}$ & $\inc_{\mathit{at}}^{\mathit{ent}}$ & \#flagged$_{\mathit{\textsc{Con}}}$\\
        \midrule
        3 atoms, $\leq$ 5 formulas&7.0&2.11&3.0&1.77& 181\\
        3 atoms, $\leq$ 10 formulas&16.0&3.42&3.0&1.29&88\\
        3 atoms, $\leq$ 15 formulas&51.0&4.39&3.0&1.04&64\\
        6 atoms, $\leq$ 5 formulas&5.0&1.82&6.0&2.33&242\\
        6 atoms, $\leq$ 10 formulas&19.0&3.17&6.0&2.50&135\\
        6 atoms, $\leq$ 15 formulas&36.0&4.28&6.0&2.27&72\\
        9 atoms, $\leq$ 5 formulas&6.0&1.65&9.0&2.33&304\\
        9 atoms, $\leq$ 10 formulas&19.0&2.92&9.0&3.08&146\\
        9 atoms, $\leq$ 15 formulas&33.0&3.99&9.0&2.92&119\\
        \bottomrule          
    \end{tabular}
    }
    \caption{Characteristics of the generated data sets over $\incat, \incmi$, with the respective \emph{max-} and \emph{entropy-} value (${\mathit{min}}$ = 0 in all cases), and $\#$ of knowledge bases that could correctly be flagged via the \textsf{Consistency} heuristic.}
    \label{tab:characteristics}
\end{table}

\subsection{Symbolic Rules/Heuristics}\label{subsec:rules}

As a basis for our approach, the machine learning models will be trained on the knowledge base data itself. However, we are also interested in how adding (symbolic) domain knowledge can improve the predictions. For this, we revisit the following rationality postulates.

\begin{description}

\item[Upper Bound.] The first symbolic rule is in regard to the so-called expressivity\footnote{Let $\allkbs^\atoms(n)=\{\kb\in\allkbs \mid |\atoms(\kb)|\leq n\}$ be the set of all knowledge bases with at most $n$ atoms, then, the expressivity $e$ of an inconsistency measure $\inc$ w.r.t. a number of atoms $n$ is defined as $e(\inc,n) = |\{\inc(\kb) | \kb\in \allkbs^\atoms(n)\}|$.} of the considered inconsistency measures \cite{thimm2016expressivity}. To recall, the range of values that can be returned by $\incat$ is limited by the number of atoms. So any prediction should never exceed this value. For the feature vector $V_\kb$, we, therefore, add a feature ``\textsf{upper-bound-x}", where \textsf{x} is the number of distinct atoms in $\kb$ and the \textsf{upper-bound} indicates the correct inconsistency degree value w.r.t. $\incat$ is bounded by \textsf{x}. The goal of this rule is to see whether the machine learning-based approaches can leverage this insight to penalize all predictions $>\textsf{x}$. Note that for the $\incmi$ measure, the number of values that can be attained w.r.t. the number of formulas grows binomial \cite{thimm2016expressivity}, so the upper-bound flag is likely not useful in vectors for $\incmi$ (and we only use it for $\incat$).

\item[Consistency (Heuristic).] The second symbolic rule is about \emph{consistent} knowledge bases: If $\kb$ is consistent, any inconsistency degree value, or prediction, must be 0 (both for $\incmi$ and $\incat$). For generating domain knowledge on consistency, a core problem is that we do not want to perform any form of (computationally expensive) solving. So the general information on whether an input knowledge base is consistent is not available in our setting. However, for the considered propositional logic, we can apply the following heuristic to correctly flag at least some consistent knowledge bases: Let $\kb$ be a knowledge base, then by definition, a set of formulas $\subseteq \kb$ is inconsistent if its smallest closed set of literals contains $a, \neg a$ for an atom. Thus, if for any $\kb$, we do not have any atom $a$ which is present in simple and negated form, $\kb$ cannot be inconsistent (For example, we can infer that $\{a, \neg b\wedge c, d\}$ cannot be inconsistent as it does not contain an atom present in both forms as described). So for the feature vector $V_\kb$, we add a feature ``\textsf{consistent}", which is true if $\kb$ does not contain a pair of literals $a, \neg a$ over all atoms as described. In this case, the correct inconsistency degree value must be 0. The goal of this rule is to see whether the approaches can leverage this insight to penalize all predictions $>0$ in this case. 
\end{description}

\subsection{Training Configuration}

We performed a ten-fold cross-validation with random shuffling for each data set and machine learning approach. We further split the training set of each fold into a validation set having the same size as the test set and a subtraining set including the remaining instances of the training set to perform a hyperparameter optimization in the form of a grid search. In particular, we fitted multiple machine learning models on the training set, selected the model with the best validation mean absolute error (MAE), and applied this model to the test set. The hyperparameters used in the grid search can be found in the appendix.

To measure the prediction performance of the machine learning models, we first calculated an MAE value for each test set, and subsequently the average and the standard deviation across all of the MAE values. We selected the MAE because our machine learning models address a regression task and it is commonly used for this type of task.   

For the MLPs, which are neural networks, we addressed the common problem of overfitting the training criterion in four regards. First, we integrated a weight decay regularization into the training procedure of the MLPs. Second, we applied early stopping in the training procedure of the MLPs, with a patience of ten epochs and the validation MAE as the stopping criterion. Third, we randomly dropped out 20\% of neurons of the first two dense layers during training of the MLPs. Fourth, we performed a ten-fold cross-validation and repeatedly tested the models' ability to generalize from the training data to unseen data. 

%
%
\section{Results}\label{sec:results}

\subsection{Experiment Results}\label{sec:exp_results}

\begin{table*}[t]
\centering
\resizebox{\textwidth}{!}{%
\begin{tabular}{@{}lrrrrrrrrr@{}}
\toprule
 & \multicolumn{9}{c}{\textbf{Data Sets}} \\ 
  \cmidrule{2-10}
\textbf{ML Models} & \begin{tabular}[c]{@{}r@{}} $\leq$ 3 atoms, \\ $\leq$ 5 formula\end{tabular} & \begin{tabular}[c]{@{}r@{}} $\leq$ 6 atoms, \\ $\leq$ 5 formulas\end{tabular} & \begin{tabular}[c]{@{}r@{}}$\leq$ 9 atoms, \\ $\leq$ 5 formulas\end{tabular} & \begin{tabular}[c]{@{}r@{}} $\leq$ 3 atoms, \\ $\leq$ 10 formula\end{tabular} & \begin{tabular}[c]{@{}r@{}} $\leq$ 6 atoms, \\ $\leq$ 10 formulas\end{tabular} & \begin{tabular}[c]{@{}r@{}} $\leq$ 9 atoms, \\ $\leq$ 10 formulas\end{tabular} & \begin{tabular}[c]{@{}r@{}} $\leq$ 3 atoms, \\ $\leq$ 15 formula\end{tabular} & \begin{tabular}[c]{@{}r@{}} $\leq$ 6 atoms, \\ $\leq$ 15 formulas\end{tabular} & \begin{tabular}[c]{@{}r@{}} $\leq$ 9 atoms, \\ $\leq$ 15 formulas\end{tabular} \\ \midrule
LR (without flags) & 11,268.685 ($\pm$4,638.938) & 4,715.627 ($\pm$2,284.037) & 4,613.901 ($\pm$2,560.487) & 4,023.273 ($\pm$1,911.731) & 2,247.905 ($\pm$1,837.624) & 814.620 ($\pm$1090.367) & 4,379.255 ($\pm$3,423.175) & 0.982 ($\pm$.056) & 0.506 ($\pm$.029) \\
LR (with flags) & 3,494.763 ($\pm$1,407.522) & 519.816 ($\pm$.325.417) & 253.612 ($\pm$254.455) & 3,789.743 ($\pm$2,214.527) & 261.589 ($\pm$313.506) & 0.573 ($\pm$.044) & 4,860.64 ($\pm$2,613.110) & 0.602 ($\pm$.054) & 0.482 ($\pm$.028) \\
Ridge (without flags) & 0.803 ($\pm$.063) & 0.935 ($\pm$.051) & 0.891 ($\pm$.088) & 0.746 ($\pm$.042) & 0.797 ($\pm$.041) & 0.854 ($\pm$.035) & 0.707 ($\pm$.067) & 0.727 ($\pm$.051) & 0.493 ($\pm$.034) \\
Ridge (with flags) & 0.588 ($\pm$.033) & 0.643 ($\pm$.032) & 0.627 ($\pm$.039) & 0.548 ($\pm$.037) & 0.543 ($\pm$.032) & 0.550 ($\pm$.037) & 0.505 ($\pm$.042) & 0.519 ($\pm$.048) & 0.456 ($\pm$.031) \\
Lasso (without flags) & 0.820 ($\pm$.051) & 0.939 ($\pm$.052) & 0.880 ($\pm$.090) & 0.757 ($\pm$.041) & 0.798 ($\pm$.061) & 0.868 ($\pm$.061) & 0.742 ($\pm$.081) & 0.775 ($\pm$.045) & 0.462 ($\pm$.037) \\
Lasso (with flags) & 0.590 ($\pm$.033) & 0.596 ($\pm$.057) & 0.589 ($\pm$.055) & 0.517 ($\pm$.052) & 0.545 ($\pm$.036) & 0.539 ($\pm$.043) & 0.491 ($\pm$.042) & 0.513 ($\pm$.048) & 0.435 ($\pm$.033) \\
MLP (without flags) & 0.831 ($\pm$.094) & 0.940 ($\pm$.053) & 0.949 ($\pm$.077) & 0.650 ($\pm$.088) & 0.772 ($\pm$.048) & 0.828 ($\pm$.050) & 0.545 ($\pm$.100) & 0.721 ($\pm$.040) & 0.460 ($\pm$.034) \\
MLP (with flags) & 0.535 ($\pm$.049) & 0.558 ($\pm$.048) & \textbf{0.547} ($\pm$.043) & \textbf{0.411} ($\pm$.062) & \textbf{0.503} ($\pm$.040) & 0.539 ($\pm$.050) & 0.345 ($\pm$.055) & 0.468 ($\pm$.052) & \textbf{0.434} ($\pm$.035) \\
\begin{tabular}[c]{@{}l@{}}MLP (with flags+constraints)\end{tabular} & \textbf{0.533} ($\pm$.041) & \textbf{0.571} ($\pm$.043) & 0.555 ($\pm$.049) & 0.412 ($\pm$.059) & 0.509 ($\pm$.043) & \textbf{0.537} ($\pm$.048) & \textbf{0.329} ($\pm$.041) & \textbf{0.467} ($\pm$.050) & \textbf{0.434} ($\pm$.037) \\ \bottomrule
\end{tabular}
}
\caption{MAE for the AT measure (\incat) with consistency and bound flag (avg. and st. deviation over ten folds, best values are marked bold).}
\label{tab:results_at_measure}
\end{table*}

\begin{table*}[ht]
\centering
\resizebox{\textwidth}{!}{%
\begin{tabular}{@{}lrrrrrrrrr@{}}
\toprule
 & \multicolumn{9}{c}{\textbf{Data Sets}} \\ 
 \cmidrule{2-10}
\textbf{ML Models} & \begin{tabular}[c]{@{}r@{}} $\leq$ 3 atoms, \\ $\leq$ 5 formula\end{tabular} & \begin{tabular}[c]{@{}r@{}} $\leq$ 6 atoms, \\ $\leq$ 5 formulas\end{tabular} & \begin{tabular}[c]{@{}r@{}} $\leq$ 9 atoms, \\ $\leq$ 5 formulas\end{tabular} & \begin{tabular}[c]{@{}r@{}} $\leq$ 3 atoms, \\ $\leq$ 10 formula\end{tabular} & \begin{tabular}[c]{@{}r@{}} $\leq$ 6 atoms, \\ $\leq$ 10 formulas\end{tabular} & \begin{tabular}[c]{@{}r@{}} $\leq$ 9 atoms, \\ $\leq$ 10 formulas\end{tabular} & \begin{tabular}[c]{@{}r@{}} $\leq$ 3 atoms, \\ $\leq$ 15 formula\end{tabular} & \begin{tabular}[c]{@{}r@{}} $\leq$ 6 atoms, \\ $\leq$ 15 formulas\end{tabular} & \begin{tabular}[c]{@{}r@{}} $\leq$ 9 atoms, \\ $\leq$ 15 formulas\end{tabular} \\ \midrule
LR (without flags) & 9,648.103 ($\pm$2,567.872) & 4,895.722 ($\pm$2,099.998) & 4,613.901 ($\pm$2,560.487) & 3,677.676 ($\pm$1,944.434) & 2,247.905 ($\pm$1,837.624) & 814.62 ($\pm$1,090.367) & 4,379.255 ($\pm$3,423.175) & 0.982 ($\pm$.056) & 0.794 ($\pm$.053) \\
LR (with flags) & 5,645.609 ($\pm$1,330.269) & 370.18 ($\pm$.744.209) & 253.612 ($\pm$254.455) & 1,915.976 ($\pm$955.207) & 261.589 ($\pm$313.506) & 0.573 ($\pm$0.044) & 4,860.64 ($\pm$2,613.11) & 0.602 ($\pm$.054) & 0.648 ($\pm$.055) \\
Ridge (without flags) & 0.737 ($\pm$.062) & 0.901 ($\pm$.056) & 0.891 ($\pm$.088) & 0.528 ($\pm$.042) & 0.797 ($\pm$.041) & 0.854 ($\pm$.035) & 0.707 ($\pm$.067) & 0.727 ($\pm$.051) & 0.696 ($\pm$.053) \\
Ridge (with flags) & 0.551 ($\pm$.051) & 0.587 ($\pm$.057) & 0.627 ($\pm$.039) & 0.468 ($\pm$.042) & 0.543 ($\pm$.032) & 0.550 ($\pm$.037) & 0.505 ($\pm$.042) & 0.519 ($\pm$.042) & 0.602 ($\pm$.050) \\
Lasso (without flags) & 0.757 ($\pm$.063) & 0.908 ($\pm$.058) & 0.880 ($\pm$.090) & 0.599 ($\pm$.041) & 0.798 ($\pm$.061) & 0.868 ($\pm$.061) & 0.742 ($\pm$.081) & 0.755 ($\pm$.045) & 0.713 ($\pm$.051) \\
Lasso (with flags) & 0.572 ($\pm$.048) & 0.567 ($\pm$.064) & 0.589 ($\pm$.055) & 0.513 ($\pm$.052) & 0.545 ($\pm$.036) & 0.539 ($\pm$.043) & 0.491 ($\pm$.042) & 0.513 ($\pm$.048) & 0.633 ($\pm$.052) \\
MLP (without flags) & 0.721 ($\pm$.066) & 0.909 ($\pm$.066) & 0.949 ($\pm$.077) & 0.508 ($\pm$.035) & 0.772 ($\pm$.048) & 0.828 ($\pm$.050) & 0.545 ($\pm$.100) & 0.721 ($\pm$.040) & 0.680 ($\pm$.056) \\
MLP (with flags) & 0.523 ($\pm$.050) & \textbf{0.529} ($\pm$.058) & \textbf{0.547} ($\pm$.043) & \textbf{0.456} ($\pm$.042) & \textbf{0.503} ($\pm$.040) & 0.539 ($\pm$.050) & 0.345 ($\pm$.055) & \textbf{0.493} ($\pm$.019) & 0.583 ($\pm$.045) \\
MLP (with flags+constraints) & \textbf{0.522} ($\pm$.043) & 0.537 ($\pm$.057) & 0.555 ($\pm$.049) & 0.459 ($\pm$.052) & 0.509 ($\pm$.043) & \textbf{0.537} ($\pm$.048) & \textbf{0.329} ($\pm$.041) & 0.495 ($\pm$.020) & \textbf{0.582} ($\pm$.048) \\ \bottomrule
\end{tabular}%
}
\caption{MAE for the MI measure (\incmi) with consistency flag (average and standard deviation over ten folds, best values are marked bold).}
\label{tab:results_mi_measure}
\end{table*}

The experiment results are shown in tables \ref{tab:results_at_measure} and \ref{tab:results_mi_measure}. The MAE indicates that the models can produce good predictions for the intended use case. Take for example the $\incmi$ measure and the dataset for the setting ``3 atoms/15 formulas": For this dataset, the actual ground truth values of $\incmi$ ranged between 0 - 51. Here, having a (predicted) value that deviates on average only 0.547 from the ground truth value may be a valuable approximation in many settings, for example, for ranking inconsistencies by their severity.  

The results for both measures further indicate that the neural network architecture (MLP) achieves the best MAE values in all settings. This shows that the MLPs are able to detect non-linearities that exist in the data and are therefore more suitable than the baseline regressions in our setting.

\begin{figure}[H]
    \centering
\includegraphics[width=0.47\textwidth]{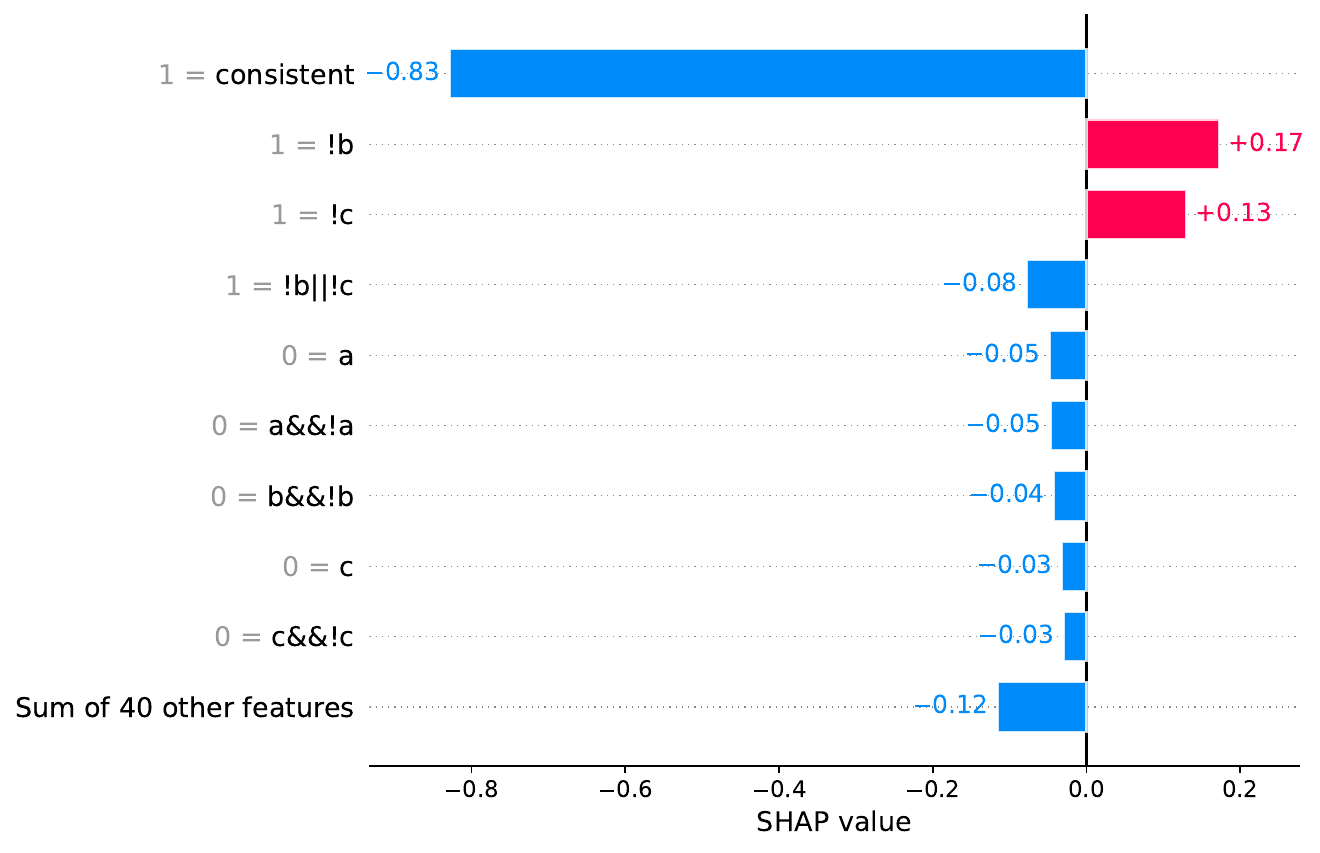}
    \caption{SHAP plot for an exemplary instance of the data set with max. 3 atoms and max. 5 formulas (KB=$\{\neg b, \neg c, \neg b\vee\neg c\}$) and the MLP with flags and constraints. The SHAP values represent the importance of the features of this instance on the predicted MI measure.}
    \label{fig:shap_plot}
\end{figure}

For both measures, all MLP models \emph{with} flags outperform their MLP counterparts without flags. This is a key outcome of our experiments, as it shows that incorporating domain knowledge in the form of symbolic constraints can indeed improve prediction performance. This is exemplified in Figure \ref{fig:shap_plot}, which shows Shapley Additive eXplanation Values (SHAP) for an exemplary prediction instance where symbolic knowledge was added in the feature vector. 

The knowledge base instance predicted for Figure \ref{fig:shap_plot} was $\{\neg b, \neg c, \neg b\vee\neg c\}$, which is consistent (and the corresponding ``consistent" flag could be set for the example). The shown SHAP values can be read s.t. the concrete values indicate the impact of the respective feature on the prediction outcome \cite{molnar2020interpretable}. As can be seen, the feature corresponding to the ``consistent" flag (set to 1 for the instance) has by far the highest impact on the prediction outcome. The model thus recognized this flag as an important feature for the prediction.

We continue with results about the experiment runtimes. 

\begin{figure}[H]
    \centering
    \includegraphics[width=0.47\textwidth]{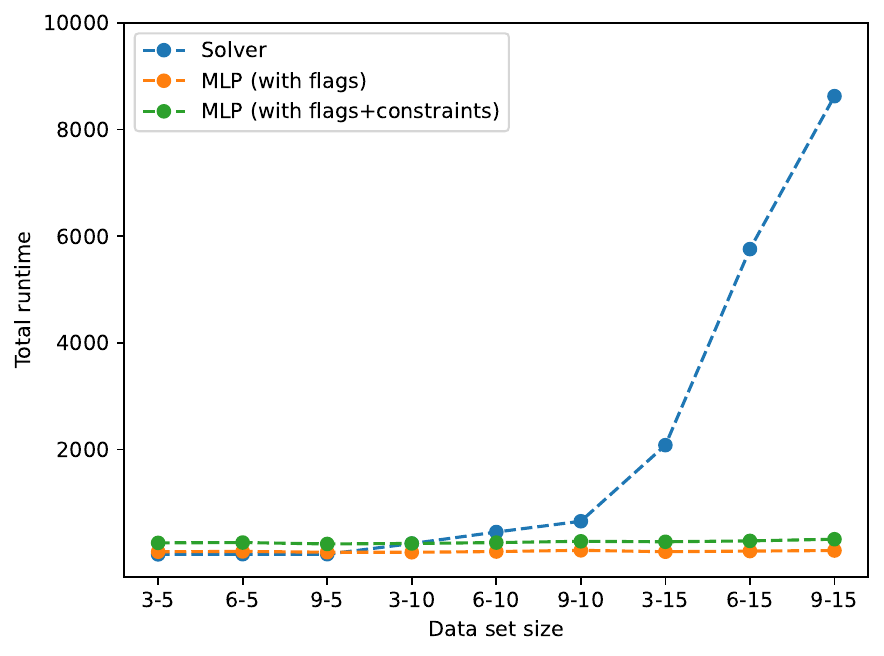}
    \caption{Runtimes (s) of the \textsf{Tweety}-library (``solver") and the MLP approaches from this work for producing inconsistency values for all data sets (AT measure). The x-axis shows the individual data sets, for example, ``3-5" means max. 3 atoms, max. 5 formulas.}
    \label{fig:runtimes}
\end{figure}

Recalling one of the main motivations of this work, a problem of computing the inconsistency measure values conventionally---especially in the envisaged continuous setting from the introduction---is that it may be computationally expensive. This is further underlined by the results shown in Figure~\ref{fig:runtimes}.
Figure \ref{fig:runtimes} shows the runtimes for computing the $\incat$ values with a conventional solver vs. obtaining the values by i) training an MLP model and then ii) using it for predictions. To clarify, as a conventional solver, we used the \textsf{Tweety} library architecture and integrated the solver presented in \cite{kuhlmann2023computing} (state-of-the-art). The results for $\incmi$ were similar and are not reported due to space reasons (using the solver from \cite{liffiton2016fast}).

The x-axis shows the different data sets and the y-axis shows the respective runtimes. For the MLP models (orange/green), it can be seen that the training time is on a constant level and is very stable. On the contrary, the runtimes for computing the degree values with a conventional solver (blue) grow exponentially with increasing data complexity. Intuitively, for smaller data sets, a conventional solver-based approach is still feasible, but Figure~\ref{fig:runtimes} clearly shows a break-even point for when the costs of using a solver outweigh the fixed costs of the machine learning-based approaches. So, especially for the envisaged continuous setting, it can be expected that the use of solver-based approaches might be unfeasible. This might then be a good use case for learning-based approximations. 

\subsection{Discussion}

Based on our experiments, the machine learning approaches could successfully be applied to predict inconsistent measure values for the analyzed data sets. In this section, we discuss the generalizability of our results.

A general problem for machine-learning-based approaches is that of overfitting. In our experiments, we addressed this problem in three regards. First, we integrated a weight decay regularization into the training procedure of the MLPs. Second, we applied early stopping in the training procedure of the MLPs, with a patience of ten epochs and the validation MAE as the stopping criterion. This ensures that the models stop training before they learn data-specific patterns too well. 
Third, we performed a ten-fold cross-validation and repeatedly tested the models' ability to generalize from the training data to unseen data.


A limitation (by design) is that labeled training data needs to be available. However, for the envisaged setting, there is a need to continuously compute inconsistency degree values. So it seems plausible that a couple of instances have to be solved anyway first, which can then be leveraged with our approach. Intuitively, for such an ML-based approach, one has to pay attention to issues such as distribution shift. Related to this, note that it is not necessary to know all distinct formulas in advance (e.g., this was exactly the case in all experiments, where some formulas from the test set were not in the training set). These formulas are fitted as new features, and the scalability results from Section \ref{sec:scalability} show how (well) the considered models could work with this. In cases of heavy distribution shift over time, naturally, it might at some point be necessary to retrain models. Determining this is however beyond the scope of this paper. The field of lifelong machine learning (or continual or incremental machine learning) provides approaches to relax this limitation by further training and adapting machine learning models over time.


Finally, regarding the considered learning approaches in general, a limitation is that for regressions one assumption is that the relationship between the independent variables and the dependent variable is linear, that is, expanding infinitely. In our setting, however, some of the target variables are bounded, and in general, all target values are non-negative. So the assumptions of this model type might not be aligned with our use cases. Nevertheless, in this context, we demonstrated that other model types with matching assumptions can be successfully applied.

In our experiments, we found that a careful definition and integration of knowledge into the models is necessary to improve the prediction performance. In particular, for the MLPs, we integrated knowledge as flags in the feature vector and constraints in the loss function used in model training. According to our results, models with only flags in the feature vector and models with both flags in the feature vector and constraints in the loss function outperform all models without flags. However, the results can be different in other domains. Thus, it is important to decide how knowledge is defined and integrated into the models \cite{besold2017neural}.

In future research, similar experiments should be repeated for inconsistency measures based on completely different principles, such as distance-based measures or semantic measures \cite{thimm2019inconsistency}.

%
%
\section{Scalability}
\label{sec:scalability}

\begin{figure}[ht]
    \centering
    \includegraphics[width=0.47\textwidth]{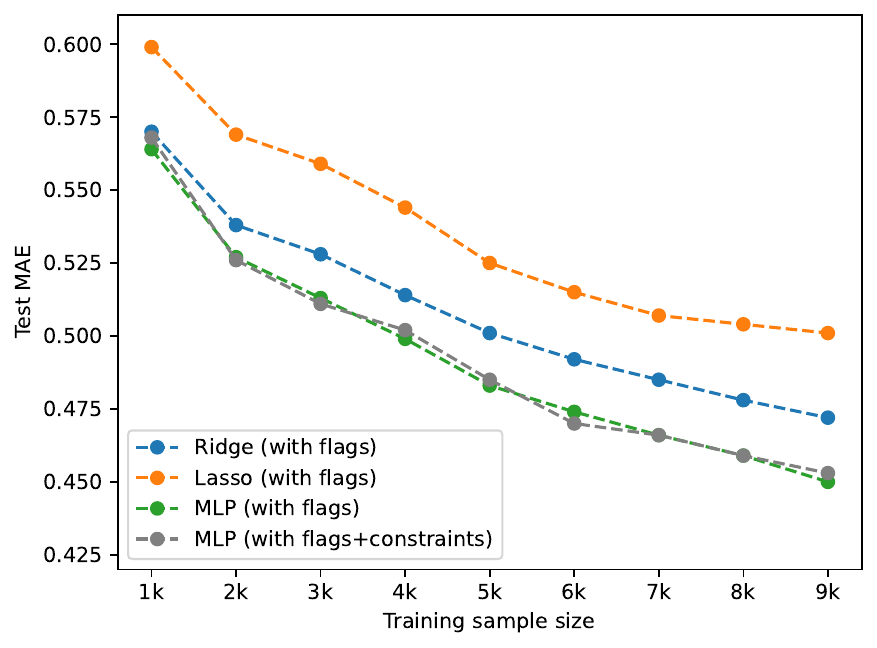}
    \caption{MAE of a data set (max. 6 atoms, max. 10 formulas) for the MI measure (\incmi) with varying training set sizes from 1K to 9K instances (average over ten folds).}
    \label{fig:scalability}
\end{figure}

We performed additional experiments to test the scalability of the machine learning approaches presented in this work. As scalability, we understand the ability of machine learning algorithms to learn more accurate models from (and generally handle) higher amounts of data. In our case, an issue we wanted to investigate in particular was the models ability to handle (larger amounts of) previously unseen formulas/features.   
For this, we conducted the following experiment. Figure~\ref{fig:scalability} shows the average MAE values for the machine learning approaches with flags and constraints when the size of the training set is extended from 1,000 to 9,000 samples.
We can see that all approaches yield improvements in the prediction performance up to the largest training set of 9,000 instances.
%
In general, all model types did not face any technical issues when dealing with the larger feature vectors and higher number of (potentially unseen) distinct formulas in the larger data sets.

%
%
\section{Related Work and Conclusion}
\label{sec:conclusion}

Our work is related to approaches applying (sub-symbolic) learning approaches to symbolic domains. In particular, our approach applies machine learning to the field of inconsistency measurement.

Most related works concerning machine learning and propositional logic are on the level of satisfiability checking. For example, works such as \cite{cameron2020predicting,bunz2017graph} present approaches for using graph neural networks to predict satisfiability. In this work, we focus on the \emph{measurement} of inconsistency, which goes beyond the binary satisfiability notion and aims to provide more fine-grained insights. In this context, there are some recent approaches that apply machine learning-based techniques to predict unsatisfiable cores in propositional logic \cite{van2020towards,selsam2019guiding,shirokikh2023machine}, which are referred to as minimal inconsistent subsets in this work. While such approaches could be used to compute degree values for some inconsistency measures (those based on $\MI$s), note that as stated many other measures are not based on unsatisfiable cores. So the problem and architecture addressed in this work are more general in that it allows to provide target values for arbitrary inconsistency measures in the scope of model training. To the best of our knowledge, this work is the first to investigate the problem of predicting degree values for inconsistency measures.

From a machine learning perspective, related works in the area of ML4KR focus on deep learning architectures. So our approach seems in line with the architectures applied in related works. It is notable here that graph neural networks have been very successfully applied to represent propositional logic for (predicting) satisfiability problems \cite{cameron2020predicting,bunz2017graph}. Therefore, it seems promising to also try to use such neural network architectures for the problem of inconsistency measurement, which we will investigate in future works. One factor in which our work is however very much aligned with related works (regardless of the architecture) is that there seems a consensus that combining symbolic constraints into the (various) deep-learning architectures seems advisable in many concrete use-cases.

The inherent complexity of computing inconsistency measures has motivated various recent works to devise novel algorithmic approaches \cite{kuhlmann2021algorithms,kuhlmann2023computing}. In this context, this work proposes a learning-based architecture. As a central distinction, the machine learning-based approach allows --- after training --- to obtain approximations in constant time. We argue there might well be various use cases where approximations are sufficient, especially in the envisaged continuous setting presented in the Introduction. Our empirical evidence shows that we could reach a clear break-even point as to where the machine learning-based approach (including training) can be performed in a faster time than with a conventional solver (see Figure \ref{fig:runtimes}).

For our concrete approach, we applied a binary encoding for representing knowledge bases. In the case of larger knowledge bases, this might lead to large feature vectors. From Section~\ref{sec:scalability} (Scalability), we could not identify any inherent technical difficulties in handling larger feature vectors. Still, in future work, we aim to investigate additional means for feature reduction that would allow us to reduce vector dimensionality. Also, different encoding forms, such as encoding knowledge with some form of abstraction, or graph neural networks, could be investigated.


Finally, while the problem investigated in this work was framed as a regression problem, a related interesting problem could be to classify two knowledge bases in which one is more inconsistent than the other. While this will not offer granular insights as in this work, there could still exist use cases where this insight would be useful, and it might be solvable with a simpler classification setup. 

In general, it seems the areas of machine learning and inconsistency measurement can be well aligned and may provide further opportunities for future work.

\section*{Appendix}
\label{sec:appendix}

A grid search was performed to tune the hyperparameters of the machine learning models. Table \ref{tab:hpo} summarizes the hyperparameters used in the grid search.

\begin{table}[ht]
\begin{center}
\caption{Hyperparameters used in grid search.}
\label{tab:hpo}
\resizebox{0.47\textwidth}{!}{
\begin{tabular}{p{2.2cm}p{2.2cm}p{4cm}}
\toprule
\textbf{ML approach} & \textbf{Hyperparameter} & \textbf{Hyperparameter range} \\ 
\midrule
\begin{tabular}[c]{@{}l@{}}Ridge \\ regression\end{tabular} & Regularization strength & $10^{-5}$, $10^{-4}$, $10^{3}$,\dots, $10^{0}$ \dots, $10^{+3}$, $10^{+4}$\\
\begin{tabular}[c]{@{}l@{}}Lasso \\ regression\end{tabular} & Regularization strength & $10^{-5}$, $10^{-4}$, $10^{3}$,\dots, $10^{0}$ \dots, $10^{+3}$, $10^{+4}$\\
MLP & Learning rate &  0.001, 0.002, 0.003\\
 & Weight decay & 0.01, 0.03, 0.05\\
& Hidden size & 32, 64, 128 \\
\bottomrule
\end{tabular}
}
\end{center}
\end{table}

\newpage
\bibliographystyle{named}
\bibliography{references}

\newpage

\end{document}